\documentclass[10pt,twocolumn,letterpaper]{article}

\usepackage{cvpr}
\usepackage{times}
\usepackage{epsfig}
\usepackage{graphicx}
\usepackage{amsmath}
\usepackage{amssymb}

\usepackage{slashbox}

\usepackage[pagebackref=true,breaklinks=true,letterpaper=true,colorlinks,bookmarks=false]{hyperref}
\usepackage{subcaption}
\usepackage{array}
\newcolumntype{L}[1]{>{\raggedright\let\newline\\\arraybackslash\hspace{0pt}}m{#1}}
\newcolumntype{C}[1]{>{\centering\let\newline\\\arraybackslash\hspace{0pt}}m{#1}}
\newcolumntype{R}[1]{>{\raggedleft\let\newline\\\arraybackslash\hspace{0pt}}m{#1}}

\cvprfinalcopy % *** Uncomment this line for the final submission

\def\cvprPaperID{32} % *** Enter the CVPR Paper ID here

\ifcvprfinal\fi

\usepackage[english]{babel}
\usepackage{fancyhdr}

\fancypagestyle{firststyle}
{
   \fancyhead[L]{Published as a conference paper at CVPR Workshop on Biometrics (2017)}
}

\begin{document}
%%%%%%%%% TITLE
\title{Deep LDA-Pruned Nets for Efficient Facial Gender Classification}
\author{Qing Tian, Tal Arbel and James J. Clark\\
Centre for Intelligent Machines \& ECE Department, McGill University\\
3480 University Street, Montréal, QC H3A 0E9, Canada\\
{\tt\small \{qtian,arbel,clark\}@cim.mcgill.ca}
\and
}

\maketitle
\thispagestyle{firststyle}

%%%%%%%%% ABSTRACT
\begin{abstract}
   Many real-time tasks, such as human-computer interaction, require fast and efficient facial gender classification. Although deep CNN nets have been very effective for a multitude of classification tasks, their high space and time demands make them impractical for personal computers and mobile devices without a powerful GPU. In this paper, we develop a 16-layer, yet lightweight, neural network which boosts efficiency while maintaining high accuracy. Our net is pruned from the VGG-16 model~\cite{simonyan2015} starting from the last convolutional (conv) layer where we find neuron activations are highly uncorrelated given the gender. Through Fisher's Linear Discriminant Analysis (LDA)~\cite{fisher1936}, we show that this high decorrelation makes it safe to discard directly last conv layer neurons with high within-class variance and low between-class variance. Combined with either Support Vector Machines (SVM) or Bayesian classification, the reduced CNNs are capable of achieving comparable (or even higher) accuracies on the LFW and CelebA datasets than the original net with fully connected layers. On LFW, only four Conv5\_3 neurons are able to maintain a comparably high recognition accuracy, which results in a reduction of total network size by a factor of 70X with a 11 fold speedup. Comparisons with a state-of-the-art pruning method~\cite{han20150} (as well as two smaller nets~\cite{krizhevsky2012,levi2015}) in terms of accuracy loss and convolutional layers pruning rate are also provided. 
\end{abstract}

%%%%%%%%% BODY TEXT
\section{Introduction}

In recent years, deep learning has revolutionized many computer vision areas due to high accuracy for a wide variety of classification tasks. Although artificial neural networks have been used for visual recognition tasks since the 1980s~\cite{lecun1989}, recent algorithms have been successful at training large networks efficiently~\cite{hinton2006,bengio2007,marc2007,hinton2012}. Given the huge amount of data that has become available, recent advances in computing have led to the emergence of deep neural nets. Even though deep learning techniques become the state-of-the-art solutions for various computer vision tasks, the requirement of a powerful GPU has made their wide deployment on general purpose PCs and mobile devices impractical. Moreover, from the `very' deep VGG-Net~\cite{simonyan2015} and GoogLeNet~\cite{szegedy2015} to the `extremely' deep Microsoft ResNet~\cite{he2015}, the competition for higher accuracy with ever larger depths is strong, rendering real-time performance on mobile devices even more out of reach.

In this paper, we explore ways to greatly prune very deep networks while maintaining or even improving on their classification accuracy. 
Our motivation stems from the current popular practice where, rather than train a deep net from scratch using all the available data, algorithm developers usually adopt a general network model and fine-tune it using a smaller dataset for the particular task. Therefore, there is a chance that some structures from the pre-trained model are not fully used for the current purpose. Our premise is that less useful structures (together with possible redundancies) could be pruned away in order to increase computational efficiency. Deep convolutional networks are generally considered to be composed of two components: the convolutional (conv) layers (alternated with activation and pooling layers) as feature extractors and fully connected (FC) layers as final classifiers \footnote{In this paper, FC layer is used in a general sense and includes all the layers after Conv5\_3.}. Deep nets outperform many traditional computer vision algorithms mainly because, given enough training data, the first component does well in learning the compositionality of the real world (by constructing very complicated features based on primitive ones).
More often than not, such features learned for a particular task are superior to handcrafted features designed with limited domain knowledge. The second component, FC layers, is essentially similar to logistic regression classifiers, which model the log-odds with a linear function. In this paper, we increase efficiency for each of the two components. We first investigate the firing patterns of last conv layer neurons through Fisher's Linear Discriminant Analysis (LDA)~\cite{fisher1936} and discover that those neuron activations are highly decorrelated for each class, which permits discarding a large number of less informative neuron dimensions without loss of information. As a result, the network complexity can be significantly reduced, which not only makes feature extraction more efficient, but also simplifies classification. In the second component, we analyze possible alternatives to the expensive FC layers for the final classification. Instead of the FC layers, which model the log-odds based on linear functions, we explore multiple alternatives such as the Bayesian classifier and SVMs (with linear and RBF kernels). 
% YOU NEED MORE DETAILS ABOUT THE EXPERIMENTS HERE
Although our approach is generally applicable to a wide range of biometrics recognition problems, we use facial gender classification as an example. Our experimental results show that when using the reduced CNN features previously extracted, both a Bayesian and SVM classifiers are able to achieve comparably high performance. They can even outperform using the original net when the dataset is particularly challenging (e.g. partial occlusions, large view changes, complex backgrounds, blurs exist). Also, the combinations of LDA-Pruned CNN nets and the Bayesian/SVM classifiers take far less space (only a few megabytes) than the original net (over 500 MB) while having a 11 times faster recognition speed. In addition, we have analyzed the relationship of accuracy change and parameters pruned away, and have compared our approach to a state of the art pruning method~\cite{han20150} as well as two smaller nets (i.e. AlexNet~\cite{krizhevsky2012} and~\cite{levi2015}).
% What do you want to say about this? NEED TO SAY SOMETHING ABOUT RESULTS HERE
According to the results, our Fisher LDA based pruning enjoys a lower accuracy loss than~\cite{han20150}, especially when the conv layers' pruning rate is high (say above 85\%). Furthermore, unlike~\cite{han20150}, our pruning approach can directly lead to space and time savings. 
The comparison with~\cite{krizhevsky2012,levi2015} justifies the superiority of pruning a deeper net over training one of smaller depth. 
The remainder of the paper is structured as follows: the relevant literature is reviewed in Section~\ref{relatedworksection}. In Section~\ref{ourmethodsection}, our light weight deep networks along with alternative classifiers are introduced. Section~\ref{experimentsection} describes our experimental validation and compares our modified nets to their originals as well as other pruned structures in terms of accuracy and efficiency. In Section~\ref{discussionsection}, our contribution and possible future directions are discussed. Section~\ref{conclusionsection} concludes the paper.

%-------------------------------------------------------------------------
\section{Related Work} \label{relatedworksection}
\subsection{Facial Gender Classification}
Gender classification from face images has long been a hot topic in biometrics research. Traditional approaches are based on hand-engineered features that can be grouped to be either global~\cite{turk1991,belhumeur1997} or local~\cite{ahonen2004,lowe1999,kumar2009}. The main problem with handcrafted features based approaches is that they require domain knowledge and may not generalize well. In this subsection, we focus on approaches that utilize features learned from neural networks.
Artificial feed-forward neural networks, for use in classification tasks, have been around for decades. In the 1990s, they began to be employed for gender classification~\cite{golomb1990, poggio1992,gutta1999}. However, the shallow structure of early neural networks has constrained their performance and applicability. It was not until late 2012 when Krizhevsky~\etal~\cite{krizhevsky2012} won the ImageNet Recognition Challenge with a ConvNet that neural networks regained attention. In the following years, various deep nets were successfully applied to a variety of visual recognition tasks including facial gender classification. Verma~\etal~\cite{verma2014} showed that the CNN filters correspond to similar features that neuroscientists identified as cues used by human beings to recognize gender. Inspired by the dropout technique in training deep nets, Eidinger~\etal~\cite{eidinger2014} trained a SVM with random dropout of some features and achieved promising results on their relatively small Adience dataset, on which Levi and Hassner~\cite{levi2015} later trained and tested a not-very-deep CNN. Instead of training on entire images, Mansanet~\etal~\cite{mansanet2016} trained relatively shallow nets using local patches and reported better accuracies than whole image based nets of similar depths. 
According to~\cite{simonyan2015}, larger depths are desired in order to gain higher accuracy. However, in general, the larger the depth, the more parameters are needed to train, and the less efficient the net will be. Therefore, it is desirable to prune deep networks to an extent that is suitable for the task and data at hand. 

\subsection{Deep Neural Networks Pruning}\label{pruningliterature}
Earlier work, targeting shallow nets, include magnitude-based biased weight decay~\cite{pratt1989}, Hessian based Optimal Brain Damage~\cite{lecun1989prune} and Optimal Brain Surgeon~\cite{hassibi1993}. More recently, aiming at deep networks, Han~\etal~\cite{han20150} developed a strategy to learn which connections are more important based on backpropagation. In~\cite{han2015}, they added two more stages of weight quantization and Huffman encoding in order to further reduce the network complexity. Their pruning is based on unit length connection, thus it may not well reflect larger scale utilities. Additionally, like other weight value based pruning methods, it assumes that large weight values represent high importance, which is not always the case (more explanations in Section~\ref{pruning}). In terms of implementation, masks are required to disregard pruned weights during network operation, which inevitably adds to the computational and storage burden. To better utilize pruning's computational advantages, Anwar~\etal~\cite{anwar2015} locate pruning candidates using particle filters in a structured way. With each pruning candidate weighted separately, the across-layer relationship is largely ignored. Last but not least, particle filters are generally expensive considering the huge number of connections in a deep net.

Unlike above weights based approaches, we treat network pruning as a dimensionality reduction problem in the feature space learned by deep nets.
The goal is not to remove dimensions of small values but rather to discard along the correct directions so that no much information will be lost. Different dimensionality reduction techniques have different measures of information.
Based on total data variance, Principal Component Analysis (PCA) has been widely used for general dimensionality reduction. However, it is not optimal in our supervised case because, without considering the labels, it may preserve unwanted variances while giving up discriminative information in low-variance dimensions.
Autoencoders~\cite{hinton2006} is also a great unsupervised approach to dimension reduction. Compared to PCA, it is able to effectively deal with complex non-linear cases. However, when looking for a subspace to project to, its aim is to preserve as much reconstruction power as possible, which is not necessarily aligned with the real utility either.
In this paper, we argue that when pruning, the information to be preserved should be task specific.
Inspirations can be drawn from a finding in neuroscience that although there are numerous neuron connections in the brain, each neuron typically receives inputs from only a small set of other neurons depending on the particular purpose~\cite{valiant2006}.

\subsection{Alternatives to Fully Connected Layers for Final Classification}
The FC layer is basically a expensive final classifier (similar to logistic regression with a computationally intensive pre-transformation process) on top of the extracted CNN features. As such, this leads to the possibility that by replacing this layer with a different classifier, a reduction in computational complexity becomes possible. Many machine learning methods, including SVM have met with some success for classification tasks, including facial gender recognition~\cite{moghaddam2002}. An advantage of SVM over logistic regression is that different kernels enable SVM to deal with linearly inseparable data. As a result, a wide variety of methods have combined neural networks and SVMs~\cite{tang2013,sharif2014,zhong2016}. However, the reasoning behind the success of such combinations is not usually provided. In~\cite{sharif2014}, Sharif~\etal combined CNN features and SVM for multiple visual recognition tasks and obtained state-of-the-art results. By replacing the softmax layer with a linear SVM and minimizing a margin-based loss, Tang~\cite{tang2013} showed a small but consistent improvement on a variety of deep learning benchmark datasets such as MNIST and CIFAR-10. Specifically for face attributes recognition, Zhong~\etal~\cite{zhong2016} found that linear SVM, together with CNN features, is able to achieve higher mean prediction accuracy than FC layers.
Another alternative is Bayesian classifier, which has a nice probabilistic interpretation similar to logistic regression but does not necessarily model the log-odds with a linear function. Due to its probabilistic nature, it may be optimal for challenging datasets with much noise and uncertainty. As demonstrated in~\cite{toews2009}, the Bayesian classifier can outperform SVM in gender recognition when there are a wide variety of occlusions and view changes present in the images.

%-------------------------------------------------------------------------
\section{Facial Gender Classification Using a Deep but Lightweight Network} \label{ourmethodsection}
\subsection{Network Structure}
In this paper, our convolutional neural network is based on the very deep VGG-16 architecture~\cite{simonyan2015} and is pre-trained using the ImageNet data in a similar way to~\cite{rothe2016}. The VGG-16 architecture is used as an example of a very deep network partly because its descendant, VGG-Face net~\cite{parkhi15}, is experimentally testified to successfully learn discriminative facial features for face verification. In our work, we fully train the network in the traditional manner before removing the FC layers, reducing the CNN feature dimensions, and plugging in alternative classifiers on top.

\subsection{Dimension Reduction in the Last Conv Layer} 
\begin{figure}[!htbp]
\centering
\begin{subfigure}{.495\linewidth}
    \includegraphics[width=\linewidth]{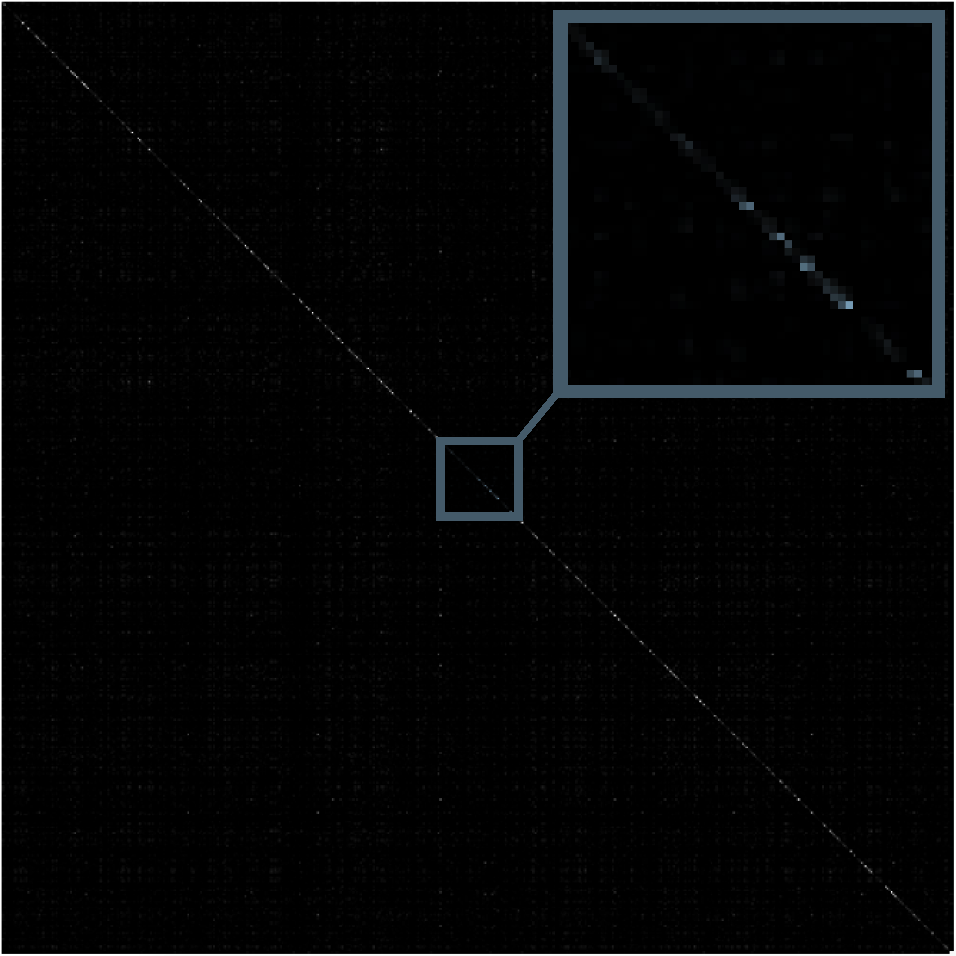}
    \caption{$S_w$ Matrix of LFW}
     \label{fig:LFWAcov}
\end{subfigure} 
\begin{subfigure}{.495\linewidth}
    \includegraphics[width=\linewidth]{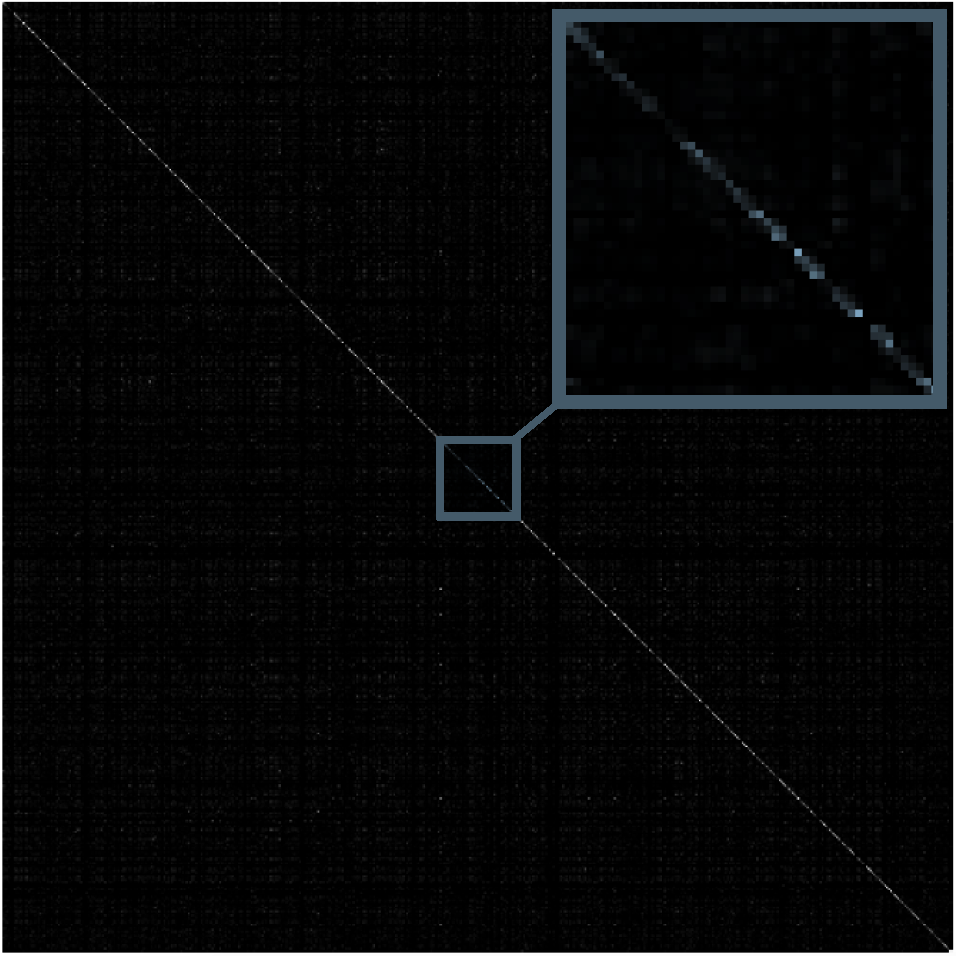}
    \caption{$S_w$ Matrix of CelebA}
    \label{fig:CelebAcov}
\end{subfigure}
\caption{$S_w$ matrices of (a) LFW and (b) CelebA. The one-pixel wide diagonals in both 512*512 matrices are so slim that they are best viewed when zoomed in (as demonstrated in the blue squares). 
%{\bf CANT SEE THIS} Not sure how to improve
}
\label{fig:sw}
\end{figure}

\begin{figure}
\centering
\includegraphics[width=.155\textwidth]{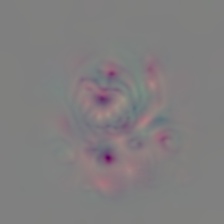}
\includegraphics[width=.155\textwidth]{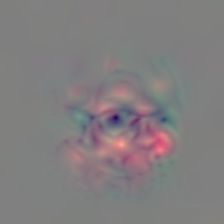}
\includegraphics[width=.155\textwidth]{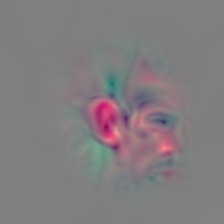}
\medskip
\includegraphics[width=.155\textwidth]{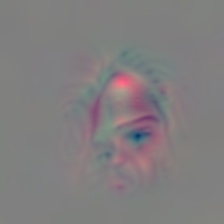}
\includegraphics[width=.155\textwidth]{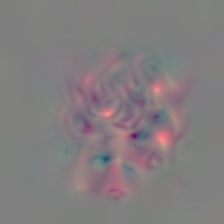}
\includegraphics[width=.155\textwidth]{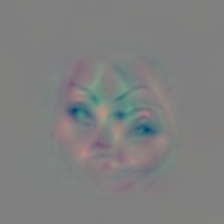}
\caption{Sample Conv5\_3 Neurons (trained On CelebA). From top left to bottom right, they fire for goatee, glasses, ear, hairline, curly hair, and noses respectively.}
\label{fig:neuronexamples}
\vskip -2mm
\end{figure}

The last conv layer is chosen as the starting point for pruning because its neurons are experimentally testified to fire more uncorrelatedly within each class than other conv layers (which, as will be seen, is critical for our LDA-based approach). Moreover, unlike FC layers, last conv layer preserves the location information and does not restrict input images to a pre-defined size or aspect ratio. In fact, many works such as~\cite{babenko2015,zhong2016} have demonstrated last conv layer's superiority over FC layers in terms of accuracy.
Layer Conv5\_3 is the last conv layer of the VGG-16 model, which has 512 neurons. 
We define the maximum activation value of a neuron as its firing score. Then for each image a 512-D firing vector can be obtained in the last conv layer, which is called a firing instance or observation. By stacking all these observations extracted from a set of images, the firing data matrix $X$ for that set is obtained. In our experiments, $X$ is normalized as a pre-processing step. 
The benefits of abandoning less useful dimensions in $X$ are twofold: 1) it compresses the data and thus has a potential for network pruning. 2) it can make the pattern hidden in the high dimensional data easier to find, which simplifies classification and possibly boost accuracy. 
As mentioned in Section~\ref{pruningliterature}, unsupervised dimensionality reduction techniques can be problematic for our case. Inspired by Fisher's Linear Discriminant Analysis~\cite{fisher1936} and its applications on face images~\cite{belhumeur1997,jain2005,bekios2011}, we adopt the intra-class correlation (ICC) to better measure information utility for gender recognition:

\begin{equation} \label{eq:icc}
ICC = \frac{s^2(b)}{s^2(b) + s^2(w)}
\end{equation}
where $s^2(w)$ is the variance within each gender, $s^2(b)$ is the variance between the two genders, and the sum of the two is the overall variance across all samples from both genders. When reducing dimensions, we are trying to maximize $ICC$, which has an equal effect of maximizing the ratio of between-gender variance to within-gender variance. The direct multivariate generalization of it is:

\begin{equation} \label{eq:interintra}
W_{opt} = \underset{W}{\arg\max} \frac{\mid W^TS_{b}W \mid}{\mid W^TS_{w}W \mid}
\end{equation}
where
\begin{equation} \label{eq:intraclassscatter}
S_{w} = \sum_{i=0:1} \sum_{x_k\in X_i}(x_k - \mu_i)(x_k - \mu_i)^T
\end{equation}
\begin{equation} \label{eq:interclassscatter}
S_{b} = \sum_{i=0:1} N_i (\mu_i - \mu)(\mu_i - \mu)^T
\end{equation}
 
\noindent and $W$ is the orthogonal transformation matrix projecting the data $X$ from its original space to a new space with the columns in $W$ being the new space's coordinate axes. $x_k$ is a firing instance of the last conv layer, $\mu$ is the mean firing vector, and $i$ indicates the gender (0 for female, 1 for male). Through analyzing $S_{w}$ for both the LFW dataset and the CelebFaces Attributes Dataset (CelebA)~\cite{liu2015} in our experiments, we find $S_{w}$ tends to be a diagonal matrix (most large values are along the diagonal and most values off the diagonal have a zero or near zero value), which is to say, the firing of different neurons in the last conv layer is highly uncorrelated given the gender.
Figure~\ref{fig:sw} shows the two $S_w$ matrices for LFW and CelebA training sets. 
These results are intuitive given the fact that higher layers capture various high-level abstractions of the data (we have also examined other conv layers, the trend is that from bottom to top the neuron activations become progressively more decorrelated). Figure~\ref{fig:neuronexamples} shows some example Conv5\_3 neuron patterns in the network trained on CelebA. Each pattern is synthesized via a regularized optimization algorithm~\cite{yosinski2015} and can be interpreted as the pattern the corresponding neuron fires most on in the input image.
Since the columns in $W$ are the (generalized) eigenvectors of $S_w$ (and $S_b$), $W$ columns are the standard basis vectors and the elements on the diagonal of $S_w$ (and $S_b$) are corresponding (generalized) eigenvalues. To maximize the ICC we simply need to select the neuron dimensions of low within-class variance and high between-class variance. For instance, although both the goatee neuron and the glasses neuron in Figure~\ref{fig:neuronexamples} have high variances (that PCA prefers), the goatee dimension has a higher chance to be selected by LDA due to its higher ICC. This corresponds to intuition, as most females do not have goatee while many males do. The direct abandonment of certain Conv5\_3 neurons greatly facilitates the pruning at all other layers.

\subsection{Pruning of the Deep Network} \label{pruning}
\begin{figure}
\begin{center}
   \includegraphics[width=\linewidth,height=0.23\textheight]{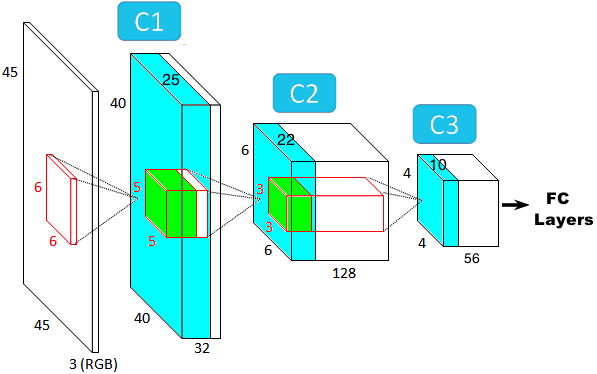}
\end{center}
\vskip -1mm
   \caption{Demonstration of pruning on filter level (cyan indicates remaining data, green represents the surviving part of a remaining next layer filter).}
\label{fig:prunedemo}
\vskip -1mm
\end{figure}

Last conv layer dimensionality reduction along neuron directions makes pruning on the neuron (filter) level possible. Instead of `masking out' smaller weights~\cite{han20150}, pruning on the neuron level directly leads to space and time savings. With the removal of a filter, the dependencies of this filter on others in previous layers are also eliminated. 
When all the dependencies on a filter from higher layers are removed, this filter can be discarded. 
Take Figure~\ref{fig:prunedemo} for example. The remaining filter outputs in a layer are colored in cyan. Corresponding useful depths of a next layer filter are colored in green (e.g. each useful $C3$ filter is represented by the small green block in column $C2$). The remaining cyan filter outputs/filters (overlapped with the green useful depths of a next layer filter) depend only on those cyan filter outputs/filters in the previous layer. Non-colored filter parts and filter outputs (filters) are thus discarded. When 106 $C2$ filters (each visualized by the small block in column $C1$) are thrown away, not only the $C2$ convolution computations with $C1$ output data are reduced by 106/128, but also $C3$ filters' depth is reduced by the same ratio (as shown in green in Column $C2$). The same applies when other layer filters are discarded. In total, 151,938 conv layer parameters are pruned away.
In our work, the dependency of a filter on others in previous layers is calculated using deconvolution (deconv)~\cite{zeiler2011,zeiler2014}, a technique mapping an max activation through lower layers all the way to the pixel level. 
As a mirrored version of the feed forward process, the deconv procedure consists of series of unpooling (utilizing stored max location switches), non-linear rectification, and reversed convolution (using a transpose of the filter).
We choose deconv over backprop for the reason that we only care about the maximum activation value of each neuron. Additionally, deconv is more robust to noise activations and vanishing gradients. It is also worth noting that unlike traditional approaches, the dependency here is learned by pooling over training samples. Its improvement over weight-based pruning is due to the fact that neural networks are non-convex and trained weights are not guaranteed to be globally optimal. Therefore, a large weight does not always indicate high importance. For example, large weights connections that have never been activated on a task specific dataset are of little use for that task. This is especially true when the network is pre-trained for a different task and we do not have enough data when fine-tuning. When pruning, the neurons with a deconv dependency smaller than a threshold is deleted. In our experiments, such a threshold is not difficult to set. Except for the first few conv layers, deconv dependencies in most other layers tend to be sparse. When the threshold is smaller than a certain value $t_0$ (e.g. when about 75\% conv parameters are pruned away in the four Conv5\_3 neurons case on LFW), an accuracy plateau is reached, beyond which point the accuracy does not change too much with the decrease of the threshold. $t_0$ is then selected as the final threshold. This guarantees no accuracy loss during the pruning process. That said, if further pruning is required, the threshold on (the highest) deconv values can be increased at the risk of sacrificing accuracy. To recover high accuracy, retraining is needed after pruning. Otherwise, the accuracy could be greatly sacrificed. To leverage the previously learned network structure (co-adapted structures and primitive features in the first few layers), the pruned networks are retrained starting from the surviving parameters without re-initializing.

\subsection{Alternative Classifiers on Top of CNN Features}

As alternatives to the expensive FC layers, SVM (with linear and RBF kernels) and Bayesian quadratic discriminant analysis are explored in our experiments based on the reduced CNN features. SVM is a deterministic, discriminative classifier, which tries to fit a hyperplane between two classes with as wide a margin as possible. It focuses on samples near the margins but does not assign attention to others. The main advantage of SVM lies in its various kernels, which, when selected properly, empower SVM to perform well even for linearly inseparable tasks. On the other hand, the Bayesian classifier is a probabilistic, generative approach. Instead of just giving a binary choice, the Bayesian classifier is able to generate a probability distribution over all (not necessarily two) classes. In cases where many sources of noise and uncertainty exist and no separating hyperplane can be easily found, the Bayesian classifier may be a better choice than SVM. That said, non-naive Bayesian quadratic discriminant analysis is vulnerable to the curse of dimensionality.

%-------------------------------------------------------------------------
\section{Experiments and Results} \label{experimentsection}
\subsection{Experimental Setup}
Our programs are implemented using Caffe~\cite{jia2014} on a Nvidia Tesla K40 GPU and a quad-core Intel i7 CPU. We modified the Caffe source code by adding modules such as filter pruning and deconv dependency calculation. 

Two datasets are used in this paper. 
The LFWA+ dataset is a richly labeled version of the popular Labeled Faces in the Wild (LFW) database~\cite{LFWTech}, originally designed for face verification tasks. It covers a large range of pose and background clutter variations. Some images even have multiple faces.
Another dataset used is the CelebFaces Attributes Dataset (CelebA)~\cite{liu2015}, which is a large-scale dataset with 202,599 images of 10,177 identities, containing the same attribute labels as in LFWA+. Despite its relatively large size, most of its images are portrait photos against simple backgrounds taken by professional photographers. 
For both databases, the train/test splits suggested in~\cite{liu2015} are adopted. All the images are pre-resized to a dimension of 224*224.

\subsection{Recognition Accuracy}

\begin{figure}[t!]
\centering 
\begin{subfigure}{0.485\linewidth}
    \includegraphics[width=\linewidth]{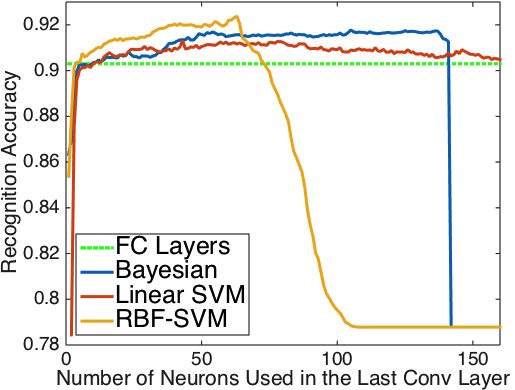}
    \caption{}
     \label{fig:LFWAresults}
\end{subfigure}
~ 
\begin{subfigure}{0.485\linewidth}
    \includegraphics[width=\linewidth]{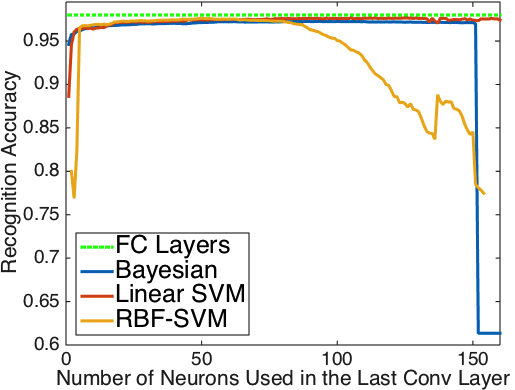}
    \caption{}
    \label{fig:CelebAresults}
\end{subfigure}
\vskip -1mm
\caption{Accuracy Comparison of Alternative Classifiers Using Pruned CNN features on (a) LFW and (b) CelebA (Blue: Bayesian, Orange: RBF-SVM, Red: Linear SVM)}
\label{fig:results}
\vskip -2mm
\end{figure}

In this section, we demonstrate and discuss the recognition results of alternative classifiers (Table~\ref{tab:highestacc}) as well as their changes with the number of preserved Conv5\_3 neurons (Figure~\ref{fig:results}).
For comparison, the results of the original deep net are also included in the first row of Table~\ref{tab:highestacc} and as a green dashed line in Figure~\ref{fig:results}.
\begin{table}
\begin{center}
\begin{tabular}{|l|l|l|}
\hline
Method & LFW & CelebA\\
\hline\hline
Original Net with FC & 90.3\% (512) & 98.0\% (512)\\
LDA-CNN+Bayesian & 91.8\% (105) & 97.3\% (94)\\
LDA-CNN+SVML & 91.3\% (43) & 97.7\% (105)\\
LDA-CNN+SVMR & 92.4\% (63) & 97.5\% (52)\\
\hline
\end{tabular}
\end{center}
\vskip -1mm
\caption{Highest recognition accuracy comparison of different approaches. SVML and SVMR represent SVM with linear and RBF kernel respectively. The accuracies reported here are the highest when a certain number (specified in the parentheses) of neurons are utilized in the last conv layer.}
\label{tab:highestacc}
\end{table}
It is worth mentioning that besides varying the number of preserved Conv5\_3 neurons, not much parameter tweaking is done. Thus, accuracies reported here are no way guaranteed to be the best. According to Table~\ref{tab:highestacc}, both the Bayesian classifier and the SVMs can achieve their highest accuracies using a small subset of last conv layer neurons on the two datasets. Particularly, the Bayesian classifier and the SVM with RBF kernel (RBF-SVM) outperformed the original net by a margin of almost 2\% on LFW. In the LFW case, RBF-SVM beats the original net when the preserved Conv5\_3 number reaches four. With more than four neurons, accuracies only improve slightly (\textless3\%) with occasional decreases. This is consistent with our hypothesis that fine-tuned deep nets possibly have many less useful and redundant structures, which may sometimes hurt accuracy. 
On CelebA, the original FC CNN has comparable (slightly higher, within 1\%) accuracies over the other LDA-CNN based classifications. In this case, most images are not as challenging as those in LFW, thus linear or generalized linear models (e.g. logistical regression based) are effectively able to separate the two classes. This can be seen as the linear SVM performs better than both RBF-SVM and the Bayesian classifier. Noticeably, the Bayesian classifier achieves a higher accuracy than linear SVM on the challenging LFW dataset where there is more uncertainty and noise. Also, on both datasets, the Bayesian classifier beats both SVMs when there are fewer than 3 neurons and has a more stable performance since it captures other information than just the margins. However, without the naive independence assumption of each dimension, the Bayesian classifier degrades drastically around 150 neurons due to the curse of dimensionality. The degradation kicks in suddenly as the space volume increases exponentially with dimensionality. Even one extra neuron dimension (e.g. from 150 to 151) can multiply the (already large) space volume. Since only a small number of neurons are needed, the Bayesian classifier is still a good choice, especially when memory resources are constrained. Although RBF-SVM performs well and attains the highest accuracy on LFW, it is slow and memory intensive to train on large datasets such as CelebA. In addition, compared to the Bayesian classifier, there are more parameters to set. Instead of choosing every parameter via cross validation, in our experiments, three sets of parameters are randomly selected for RBF-SVM and the accuracies reported here are comprised on their average output. Also, in both Figure~\ref{fig:LFWAresults} and Figure~\ref{fig:CelebAresults}, the accuracy of the RBF-SVM kernel first increases and begins to decrease suddenly due to overfitting. This occurs a little later on CelebA than on LFW because of CelebA's larger size.

Compared to Bayesian and RBF-SVM, linear SVM performed similarly to the original FC classifier on both datasets. This is intuitive in that FC layers, including softmax, is basically a logistic regression classifier with a transformed input and a linear SVM can be derived from logistic regression. Nevertheless, as will be shown in Subsection~\ref{complexitysubsection}, the LDA-CNN-SVM structure is much more efficient than the original net.

\subsection{Accuracy Change vs. Parameter Pruning Rate}
\begin{figure}
\begin{center}
   \includegraphics[width=\linewidth,height=0.31\textheight]{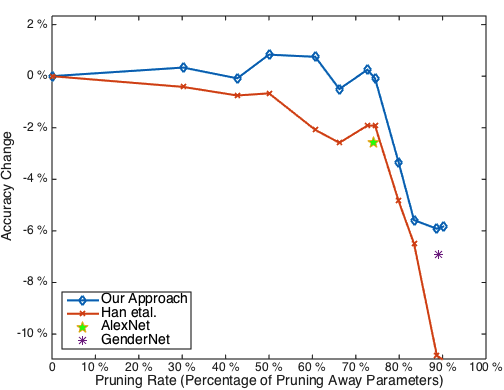}
\end{center}
\vskip 0mm
   \caption{Accuracy change vs. conv layers pruning rate. Only 4 Conv5\_3 neurons are used (when pruning rate$\neq0$). Comparisons: Han etal.\cite{han20150}, AlexNet~\cite{krizhevsky2012}, GenderNet~\cite{levi2015}.}
\label{fig:pruningcompare}
\vskip -1.5mm
\end{figure}

In this subsection, we analyze the relationship of parameters pruning rate and accuracy change and compare our results with a state-of-the-art pruning approach~\cite{han20150} as well as two smaller net structures, i.e. AlexNet~\cite{krizhevsky2012} (without filter grouping) and GenderNet~\cite{levi2015}. As shown in previous subsection, by preserving only four neurons in the last conv layer, we are able to achieve an accuracy comparable to the original DNN on the LFW dataset. We take this case as an example. Figure~\ref{fig:pruningcompare} demonstrates its accuracy change/pruning rate relationship (by varying the dependency threshold).
It is worth noting that unlike~\cite{han20150}, we only retrain both pruned networks once. Additionally, for fair comparison, we also use (pruned) FC layers to classify our Fisher LDA reduced features. Since our goal is to prune CNN features for use with alternative lightweight classifiers, we keep the pruning rate of FC layers the same for both pruning approaches and the pruning percentage reported is of only the conv layers. As a side note, all common hyper-parameters are set the same for both approaches and little tweaking of parameters is involved. That said, the batch size sometimes needs to be adjusted in order to escape local minima.
As can be seen from Figure~\ref{fig:pruningcompare}, our approach has higher accuracies across different pruning rates than~\cite{han20150}. At some points, accuracy can even improve slightly (\textless0.8\%) when pruning due to the redundant and less useful structures in the hidden layers. Also, for our approach, only about a quarter of all the conv layer weights are enough to maintain a comparable discriminating power. When pruning to around the same number of parameters as the AlexNet~\cite{krizhevsky2012}, both pruning approaches enjoy higher accuracies, which justifies the superiority of pruning pre-trained larger networks over training shallow ones. 
However, around 80\%, both approaches suffer greatly from pruning. When the pruning rate reaches 84\%,~\cite{han20150} is not able to recover itself through retraining and performs even worse than the shallow GenderNet~\cite{levi2015}. Ours, on the other hand, seems to regain stability after the drastic fall and performs better than the fixed net. Our approach's better performance mainly stems from the awareness of each neuron's contribution to the final discriminating power when pruning the net. In other words, our approach's dependency is across all layers. In contrast, the dependency in~\cite{han20150} is of length one. It may prune away small weights that contribute to more informative neurons in the last conv layer because the effects of small weights are possible to be accumulated over layers or be enlarged by large weights in other layers. Pruning away small weights in a certain layer is actually cutting off whole connections from the raw pixel level to the final classification stage. Even if weight magnitude is a good pruning measure, the importance of a whole bottom-to-top connection should not be measured by a length one weight.
In next subsection, we will provide a computational complexity analysis in terms of both space and time.

\subsection{Complexity Analysis}\label{complexitysubsection}
\begin{figure}
\begin{center}
   \includegraphics[width=\linewidth]{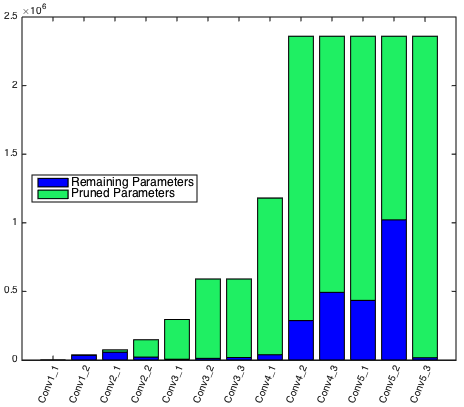}
\end{center}
\vskip -2mm
   \caption{Demonstration of layerwise structure complexity reduction by keeping the 4 discriminative Conv5\_3 neurons.}
\label{fig:layerwisepruning}
\vskip -0mm
\end{figure}

To gain more insight into our pruning method, Figure~\ref{fig:layerwisepruning} and Table~\ref{tab:speedtest} offer a detailed layerwise space and time complexity analysis.
\begin{table*}
\begin{center}
\begin{tabular}{|l|C{1.227cm}|C{1.227cm}|C{1.227cm}|C{1.227cm}|C{1.227cm}|C{1.227cm}|C{1.227cm}|C{1.227cm}|}
\hline
\backslashbox{Method}{Layer} & Conv1\_1 & Conv1\_2 & Conv2\_1 & Conv2\_2 & Conv3\_1 & Conv3\_2 & Conv3\_3 & Conv4\_1\\
\hline
Original CNN+FC Layers & 70.96 & 405.39 & 183.60 & 362.15 & 171.64 & 341.23 & 341.33 & 166.94\\
LDA-CNN+Bayesian/SVM & 18.02 &  98.27 & 39.68 & 31.96 & 3.59 & 6.43 & 9.92 & 3.79\\
\hline
Speedup Ratio & 3.93 & 4.13 & 4.63 & 11.33 & 47.83 & 53.06 & 34.41  & 44.08\\
\hline
\end{tabular}
\vskip 0.5mm
\begin{tabular}{|l|C{1.227cm}|C{1.227cm}|C{1.227cm}|C{1.227cm}|C{1.227cm}|C{0.68cm}|C{0.68cm}|C{0.68cm}|C{1.227cm}|}
\hline
\backslashbox{Method}{Layer} & Conv4\_2 & Conv4\_3 & Conv5\_1 & Conv5\_2 & Conv5\_3 & \multicolumn{3}{c|}{\begin{tabular}{@{}c@{}}FC Layers \\ \hline BC \hspace{0.8mm} SVML \hspace{0.5mm} SVMR \end{tabular}}& \textbf{Total}\\
\hline
Original CNN+FC Layers & 333.75 & 333.98 & 85.69 & 85.70 & 85.63 & \multicolumn{3}{c|}{283.20} & \textbf{3306.50}\\ \cline{7-9}
LDA-CNN+Bayesian/SVM & 18.11 & 28.07 & 6.79 & 11.92 & 0.84 & 0.04 & 0.01 & 0.05 & \textbf{286.86}\\
\hline
Speedup Ratio & 18.43 & 11.90 & 12.63 & 7.19 & 101.68 & 7E3 & 3E4 & 6E3 & \textbf{11.53}\\
\hline
\end{tabular}
\end{center}
\vskip -5mm
\caption{Per image recognition time comparison of different approaches in all layers (in milliseconds). BC is short for the Bayesian classifier, SVML and SVMR stand for SVM with linear and RBF kernel respectively. FC layers here refer to all the layers after Conv5\_3. The tests are run on the CPU.}
\label{tab:speedtest}
\vskip -4mm
\end{table*}
According to Figure~\ref{fig:layerwisepruning}, most parameters in the middle conv layers (Conv2\_2 to Conv4\_1) do not help with our task. Compared to later layers, the first three layers have relatively low reduction rates. This is easy to understand given the observation that earlier layers contain more generic features such as edge and color blob detectors that could be useful to all classes. In addition, our approach's high pruning rate can directly contribute to lower memory requirements because unlike~\cite{han20150}, it enables us to discard (rather than disregard) filter weights. In~\cite{han20150}, masks are needed in the retraining stage to freeze zero weights. As a result, besides large overhead costs of extra masks, the number of convolutional operations does not actually change. That said, if masking is also applied in testing (at the cost of even more space), time will be saved since many multiplication operations are replaced by a simpler mask checking.
Complexity can be further reduced if we replace the (pruned yet still large) FC layers with our lightweight alternatives. Since our alternative classifiers are based only on the highest activation, they are more robust to noise (no performance degradation is incurred even when the FP16 precision is used). Compared to the original deep net model of over 500 MB, our pruned model is very light and takes up only 7 MB (with no accuracy loss). For the Bayesian classifier, the storage overhead can be ignored when only four neurons are used (even when all neurons are utilized in Conv5\_3, the extra space required is just about 2 MB). For SVMs, the extra storage needed depends on the number of trained support vectors. In the LFW and four Conv5\_3 neurons case, it is only about 30KB for both SVMs. Given the fact that most of today's latest cellphone models have only 1 or 2 GB RAM, the low storage requirements of our pruned nets are critical if we want to go from off-chip to on-chip. 

Table~\ref{tab:speedtest} shows the recognition speed comparison between the original net and our pruned model. The original net is trained in the GPU mode using Caffe while tested with the CPU mode on. To avoid as much Caffe overhead as possible, we implement features extraction using survived filters ourselves utilizing all the four cores.
According to the table, our LDA-Pruned model is faster at all conv layers than the original net. Besides the last conv layer, the middle layers with high structure reduction rates also enjoy a large speedup. Nonetheless, the relation is nonlinear owing to the different dimensions of each layer's input data. In total, a 11-fold speedup is achieved by using our pruned model. It is worth noting that both the SVMs and the Bayesian classifier (based on the reduced CNN features) are significantly faster than the original FC layers in classification. The Bayesian classifier's speed is somewhere between the two SVMs.

%------------------------------------------------------------------------
\section{Discussion and Future Directions} \label{discussionsection}
While many big datasets are the property of large corporations (e.g. DeepFaces~\cite{taigman2014}), academic datasets are relatively small. Compact pruned nets like ours are easier to train and retrain, thus alleviating the data constraint to some extent and simultaneously improving on the generalizability~\cite{lecun1989prune}. 
Furthermore, due to the low space and time complexity, pruned nets can possibly be embedded on the chip SRAM to tackle real-time video streams.
Although this paper leverages VGG-16 for gender recognition, it is likely that the high decorrelation found in the last conv layer is common to some other CNNs and tasks as well. However, more tests are needed for the above to be seen. It is also appealing to train and prune deep nets for other facial traits and explore their possible shared structures. As shown in Figure~\ref{fig:neuronexamples}, when we train a deep net for gender classification, some other attributes are obtained in the last conv layer.

%------------------------------------------------------------------------
\section{Conclusion} \label{conclusionsection}
In this paper, we develop a deep but lightweight CNN that can boost efficiency while maintaining accuracy for facial gender classification. It is pruned from the VGG-16 model, whose most last conv layer neurons tend to fire uncorrelatedly within each class. Through Fisher LDA, these neurons in dimensions that have low ICC were discarded, thereby greatly pruning the network and significantly increasing efficiency. As the result, the approach can be useful in contexts where fast and accurate performance is desirable but where expensive GPUs are not available (e.g. embedded systems). 
Our LDA based pruning is better than weight value based approaches because filter weights can be large but unimportant for the specific limited task when the pre-training is done on a large dataset of general recognition purposes (e.g. ImageNet).
By combining with alternative classifiers, the approach is shown to achieve higher or comparable accuracies to the original net on the LFW and CelebA datasets, but with a reduction of model size by 70X, and with a subsequent 11-fold speedup.
%------------------------------------------------------------------------
\section*{Acknowledgment}
The authors gratefully acknowledge the support of NSERC and McGill MEDA Award. 
We would also like to acknowledge the support of NVIDIA Corporation with the donation of the Tesla K40 GPU used for this research.

{\small
\bibliographystyle{ieee}
\bibliography{egbib}
}

\end{document}